\documentclass[conference]{IEEEtran}
\IEEEoverridecommandlockouts
\usepackage{cite}
\usepackage{amsmath,amssymb,amsfonts}
\usepackage{algorithmic}
\usepackage{textcomp}
\usepackage{caption} 
\usepackage{array}
\usepackage{listings}
\usepackage{tikz}
\usepackage{booktabs}
\usepackage{tabularx}
\usetikzlibrary{arrows.meta,positioning,fit,calc}
\usepackage{xcolor}
\usepackage{tikz}
\usepackage{graphicx}
\usepackage{booktabs}
\usetikzlibrary{arrows.meta,positioning,fit,calc}
\usepackage{adjustbox}
\usepackage{float}
\usepackage[caption=false,font=footnotesize]{subfig}
\usepackage{listings}
\usepackage{tabularx}
\vspace{.25\baselineskip}
\usepackage[most]{tcolorbox}
\newtcolorbox{sbboxaccent}[1][]{colframe=phase2, colback=teal!3,
  colbacktitle=teal!10, coltitle=black, fonttitle=\bfseries,
  left=6pt,right=6pt,top=6pt,bottom=6pt, boxrule=0.7pt, sharp corners, title=#1}
\tcbset{enhanced jigsaw, breakable, boxrule=0.6pt, sharp corners}
\newtcolorbox{sbbox}[1][]{colframe=black, colback=gray!2,
  colbacktitle=gray!10, coltitle=black, fonttitle=\bfseries,
  left=6pt,right=6pt,top=6pt,bottom=6pt, title=#1}
\newcolumntype{Y}{>{\raggedright\arraybackslash}X}

\lstdefinestyle{sbcode}{
  basicstyle=\ttfamily\footnotesize,
  frame=single,
  breaklines=true,          
  breakatwhitespace=false,  
  columns=fullflexible,      
  keepspaces=true,
  showstringspaces=false
}

\definecolor{reflectOrange}{RGB}{217,95,2}
\colorlet{phase1}{blue!55!black}
\colorlet{phase2}{teal!65!black}
\colorlet{phase3}{purple!65!black}
\colorlet{phase4}{gray!60!black}

\usetikzlibrary{positioning}
\usetikzlibrary{positioning}
\usepackage[utf8]{inputenc}
\def\BibTeX{{\rm B\kern-.05em{\sc i\kern-.025em b}\kern-.08em
    T\kern-.1667em\lower.7ex\hbox{E}\kern-.125emX}}

\usepackage[hidelinks]{hyperref} 
\definecolor{linkblue}{RGB}{0,45,114}
\begin{document}

\title{ScenarioBench: Trace-Grounded Compliance Evaluation for Text-to-SQL and RAG}

\author{\IEEEauthorblockN{1\textsuperscript{st} Zahra Atf}
\IEEEauthorblockA{\textit{Faculty of Business and Information Technology} \\
\textit{Ontario Tech University}\\
Oshawa, Canada \\
ORCID: 0000-0003-0642-4341}
\and
\IEEEauthorblockN{2\textsuperscript{nd} Peter R. Lewis}
\IEEEauthorblockA{\textit{Faculty of Business and Information Technology} \\
\textit{Ontario Tech University}\\
Oshawa, Canada \\
ORCID: 0000-0003-4271-8611}
}
\IEEEaftertitletext{%
\vspace{-1.0\baselineskip}%
{\color{linkblue}\noindent\large\bfseries
This paper was accepted for presentation at the \emph{LLMs Meet Databases (LMD) Workshop}
at the 35th IEEE International Conference on Collaborative Advances in Software and Computing.\\
Workshop website: \textcolor{linkblue}{\url{https://sites.google.com/view/lmd2025/home}}%
\par\medskip}%
}

\maketitle

\begin{abstract}
\noindent
ScenarioBench is a policy-grounded, trace-aware benchmark for evaluating Text-to-SQL and retrieval-augmented generation in compliance settings. Each YAML scenario ships with a \emph{no-peek} \emph{gold-standard} package—expected decision, minimal witness trace, governing clause set, and canonical SQL—enabling end-to-end scoring of both \emph{what} a system decides and \emph{why} it decides so. Systems must justify outputs with clause IDs retrieved from the same policy canon, making explanations falsifiable and audit-ready. The evaluator reports decision accuracy, trace quality (completeness/correctness/order), retrieval effectiveness, SQL correctness via result-set equivalence, policy coverage, latency, and an explanation-hallucination rate. A normalized Scenario Difficulty Index (SDI) and its budgeted variant (SDI-R) aggregate results while pricing retrieval difficulty and time. Unlike Spider and BIRD, which assess Text-to-SQL without clause-level provenance, and unlike KILT/RAG-style setups that lack a compliance-oriented trace protocol, ScenarioBench ties decisions to clause-ID evidence under a strict grounding and no-peek discipline. On a seed, synthetic suite (N=16), decision metrics saturate (accuracy and macro-F1 = 1.000), while a single budgeted reflection step closes trace gaps (trace-completeness and policy-coverage $0.541!\rightarrow!1.000$) with zero hallucination at approximately +1 ms latency, indicating that marginal gains come from justification quality under explicit time budgets.
\end{abstract}

\section{Introduction}
Evaluations of Text-to-SQL and retrieval-augmented generation (RAG) rarely enforce \emph{trace-grounded} decisions suitable for regulated workflows, where reviewers must verify which clauses justify a decision and whether those clauses were actually retrieved by the system being scored. Text-to-SQL benchmarks such as \emph{Spider} emphasize query correctness and cross-domain generalization but do not require clause-level provenance or pricing of retrieval difficulty in end-to-end scoring \cite{Yu2018Spider}. In parallel, RAG work has improved effectiveness and discussed hallucinations, yet provenance remains under-specified and unsupported attributions persist \cite{Ji2023Hallucination, Lewis2020RAG, Wang2024RAGBestPractices}. Production stacks often mix lexical and vector search (e.g., FAISS) \cite{Johnson2017FAISS}, but lack a policy-grounded protocol that unifies Text-to-SQL accuracy, retrieval/trace quality, and compliance auditing under a single evaluator.
Concrete compliance cases make the requirements explicit. Under Canada’s Anti-Spam Legislation (CASL), commercial email must include an unsubscribe mechanism, clear sender identification, and an appropriate consent status, with exceptions for transactional messages. In such settings, audits demand SQL correctness judged by \emph{result-set equivalence on clause IDs}—not surface-form identity—so that decisions are falsifiable and contestable, and any cited evidence can be checked against the same canon the system queried.
\noindent
Recent work underscores that trust in AI is tightly coupled with explainability and the calibrated handling of uncertainty, particularly in regulated decision contexts \cite{atf2025trust}. Human-centric perspectives emphasize explanations that surface reasons, limits, and confidence in ways stakeholders can act upon \cite{atf2023humancentricity}. Complementarily, rule-governed schemes for reason-giving offer a principled way to structure uncertainty disclosures and reduce unsupported attributions \cite{atf2025rulebased}. Building on these insights, \emph{ScenarioBench} operationalizes trace-grounded evaluation—requiring systems to justify outcomes with clause-level evidence—so that explainability contributes to verifiable trust under bounded resources and domain constraints \cite{Atf2025}.
\emph{ScenarioBench} addresses this gap with three contributions. First, a \textbf{grounding invariant} requires systems to justify outputs using clause IDs they themselves retrieved from the policy canon, making explanations falsifiable and audit-ready. Second, a \textbf{dual materialization} of a jurisdiction-agnostic canon—Prolog facts for deterministic rule execution and a SQL \texttt{Policy\_DB} for retrieval and NLQ-to-SQL—preserves referential integrity across views. Third, a \textbf{difficulty- and latency-aware aggregation} combines core signals—decision accuracy/macro-F1, trace completeness/correctness/order, SQL result-set equivalence, retrieval effectiveness, and an \emph{unsupported-attribution rate}—into a normalized Scenario Difficulty Index (SDI) and a budgeted variant (SDI-R) that price retrieval difficulty and time.
We therefore ask, in a form we can evaluate end-to-end without external peeking: does clause grounding improve decision reliability over ungrounded baselines; how do retrieval choices (BM25 and hybrid) shift justification quality as measured by trace completeness/correctness and unsupported attribution; and, under a per-scenario wall-clock budget \(B_{\text{time}}\), can short, constrained reflection deliver consistent gains in explanation quality without access to the \emph{gold-standard} package?

\section{Task \& Gold-Standard Protocol}
Given a natural-language \emph{scenario} with policy-relevant \emph{context} and candidate \emph{content}, the system must output\\
(i) a discrete \emph{decision} (e.g., \texttt{approve}/\texttt{revise}/\texttt{reject}/\texttt{escalate}) and 
\\(ii) an \emph{execution trace} that \emph{explicitly grounds} the decision in retrieved policy clauses. Traces are ordered lists of $(\texttt{clause\_id},\texttt{role})$ with $\texttt{role}\in\{\texttt{applies},\texttt{exception},\texttt{precedence}\}$ and brief rationales. All identifiers in the trace must be drawn from the system’s own retrieval (lexical/vector/hybrid); citing knowledge outside the canon is disallowed.

For each YAML scenario, a \emph{gold-standard} package enables trace-grounded, end-to-end evaluation: a gold decision and a minimal sufficient witness trace, a governing clause set (necessary/sufficient identifiers, closed over exceptions/precedence), and a canonical SQL query over \texttt{Policy\_DB}. SQL correctness is judged by \emph{result-set equivalence} on \texttt{clause\_id}s rather than string identity. Optionally, a gold top-$k$ ranking (\texttt{qrels}) provides stable targets for Recall@k, MRR, and nDCG without exposing answers at inference. Gold labels are derived deterministically from the synthetic canon via a reference rule engine with stable tie-breaking. The gold-standard package is evaluator-only (\emph{no-peek}).

Scenarios compile to Prolog facts for deterministic rule execution and to a relational \texttt{Policy\_DB} with BM25 and vector indexes. A one-to-one mapping between \texttt{clause\_id}, the corresponding Prolog predicate, and \texttt{Policy\_DB.clauses.id} preserves referential integrity; a validator enforces YAML well-formedness, ID consistency, and trace–schema constraints. Two invariants govern scoring: grounding (all predicted trace IDs $\subseteq$ retrieved@\,$k$) and no-peek (systems cannot access gold decisions/traces/rankings at inference).

The evaluator reports decision accuracy and macro-F1; trace completeness, correctness, and order (Kendall-$\tau$ against the gold witness); retrieval effectiveness (Recall@k/MRR/nDCG when \texttt{qrels} exist); NLQ-to-SQL accuracy by result-set equivalence; policy coverage; latency; and an explanation–hallucination rate computed against the gold closure. A budgeted reflective loop may adapt retrieval, NLQ, and rule parameters without gold access, subject to a per-scenario wall-clock budget $B_{\text{time}}$ and early stopping on minimal improvement; all changes are logged for audit. Default retrieval settings (e.g., $k{=}10$ and $(w_{\text{vec}},w_{\text{bm25}}){=}(0.6,0.4)$) appear in Table~\ref{tab:design-defaults}.

In practice, the protocol is a constrained, auditable pipeline: synchronized SQL/Prolog views feed retrieval and NLQ-to-SQL, which in turn drive a rule engine that must return a trace grounded strictly in the system’s own top-$k$; standard logs (JSONL/CSV) capture configuration, evidence, and timing. If enabled, a short, budgeted reflection adjusts only local knobs (e.g., $k$, hybrid weights, template choice) under the same no-peek discipline.

\begin{figure}[!t]
\centering
\begin{adjustbox}{max width=\columnwidth}
\begin{tikzpicture}[
  x=1cm, y=1.2cm, >=Stealth,
  node distance=11mm and 10.5mm,
  every node/.style={font=\small\bfseries, align=center},
  box/.style={draw, rounded corners, inner sep=2pt, minimum height=12.5mm, text width=25mm},
  nbox/.style={draw, rounded corners, inner sep=2pt, minimum height=12.5mm, text width=24mm},
  gboxA/.style={draw, dashed, rounded corners, inner sep=3.5pt, line width=0.6pt, color=phase1},
  gboxB/.style={draw, dashed, rounded corners, inner sep=3.5pt, line width=0.6pt, color=phase2},
  gboxC/.style={draw, dashed, rounded corners, inner sep=3.5pt, line width=0.6pt, color=phase3},
  gboxD/.style={draw, dashed, rounded corners, inner sep=3.5pt, line width=0.6pt, color=phase4},
  flow/.style={->, semithick, color=black},
  refl/.style={->, dashed, very thick, color=reflectOrange},
  eval/.style={->, dotted, semithick, color=gray!75}
]
\node[box] (yaml) {YAML\\Scenarios};
\node[box, right=of yaml] (compiler) {Scenario\\Compiler};
\node[box, right=of compiler] (policydb) {Policy-DB\\(SQL)};
\node[box, below=of compiler] (prolog) {Prolog\\Facts};
\draw[flow] (yaml) -- (compiler);
\draw[flow] (compiler) -- (policydb);
\draw[flow] (compiler) -- (prolog);
\node[gboxA, fit=(yaml)(compiler)(policydb)(prolog),
      label={[font=\scriptsize\bfseries, text=phase1]above:{(1) Protocol \& Materialization}}] {};

\node[nbox, below=of policydb] (indexes) {Indexes:\\BM25 / Vector};
\node[nbox, right=of indexes] (nlq) {NLQ-to-SQL\\Layer};
\node[nbox, below=of indexes] (retriever) {Retriever\\(top-$k$ clauses)};
\node[nbox, below=of nlq] (rule) {Rule\\Engine};
\node[nbox, below=of rule, text width=28mm] (out) {Decision + Trace\\(grounded in retrieved clauses)};
\draw[flow] (policydb) -- (indexes);
\draw[flow] (policydb) |- (nlq);
\draw[flow] (indexes) -- (retriever);
\draw[flow] (retriever) -- (rule);
\draw[flow] (nlq) -- (rule);
\draw[flow] (rule) -- (out);
\node[gboxB, fit=(indexes)(nlq)(retriever)(rule)(out),
      label={[font=\scriptsize\bfseries, text=phase2]above:{(2) Inference}}] {};

\node[nbox, below=of out, text width=28mm] (log) {Standard Log\\(JSONL/CSV)};
\node[nbox, below=of log, text width=28mm] (metrics) {Metrics \& SDI/SDI-R:\\
Acc, SQL Acc, Recall@k, MRR, nDCG,\\
Trace C/C/O, Coverage, Latency, Hallucination};
\draw[flow] (out) -- (log);
\draw[flow] (log) -- (metrics);
\node[gboxC, fit=(log)(metrics),
      label={[font=\scriptsize\bfseries, text=phase3]below:{(3) Logging \& Evaluation}}] {};

\node[nbox, dashed, below=12mm of metrics, text width=36mm] (gold) {Gold-standard package:\\
decision, execution trace, clause set, canonical SQL\\
\scriptsize evaluator-only (no-peek)};
\draw[eval] (gold.north) -- (metrics.south);
\node[gboxD, fit=(gold),
      label={[font=\scriptsize\bfseries, text=phase4]below:{(4) Gold-standard (evaluator-only)}}] {};

\node[nbox, above=12mm of nlq, text width=36mm] (critics) {Critics:\\Retrieval \; / \; SQL \; / \; Rule \; / \; Trace};
\draw[refl] (metrics.north) to[out=80,in=220]
  node[midway, above, xshift=0.6mm,
       font=\scriptsize\bfseries, text=reflectOrange,
       fill=white, inner sep=0.9pt, rounded corners=0.6pt] {signals, $\Delta$metrics} (critics.west);
\draw[refl] (critics.south west) to[out=-90,in=120] (retriever.north);
\draw[refl] (critics.south) -- (nlq.north);
\draw[refl] (critics.south east) to[out=-90,in=60] (rule.north);

\node[font=\scriptsize\bfseries, below=1.2mm of out,
      fill=white, inner sep=1pt, rounded corners=0.8pt]
  {\emph{Grounding:} trace may cite only system-retrieved clauses.};
\node[font=\scriptsize\bfseries, above=1.2mm of critics]
  {$B_{\text{time}}$ per scenario (max allowed time); $\le$2 cycles};
\end{tikzpicture}
\end{adjustbox}
\caption{ScenarioBench pipeline with four phases. Decisions are grounded in retrieved clauses; inference respects a per-scenario latency budget $B_{\text{time}}$.}
\label{fig:scenariobench-pipeline}
\end{figure}
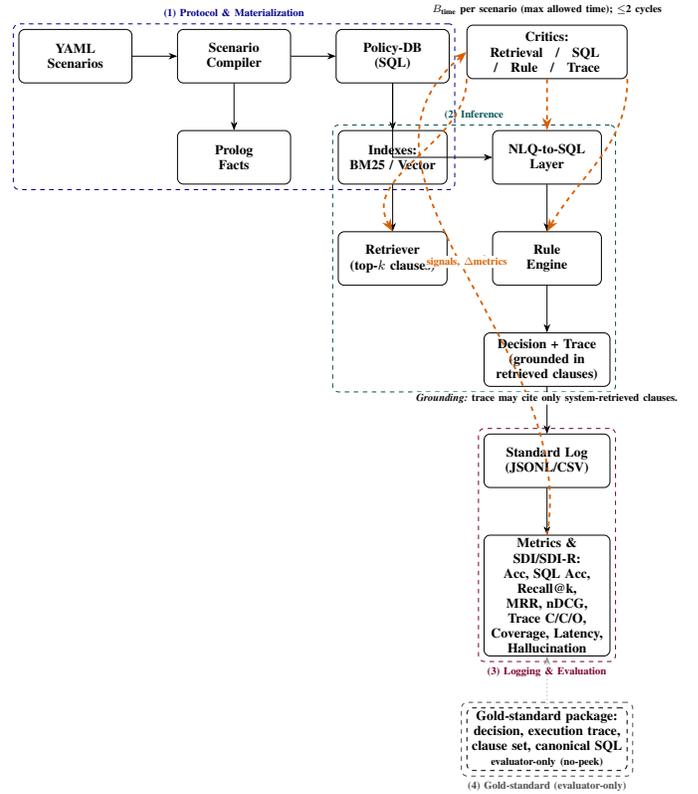

This decomposition foregrounds the grounding invariant (all trace IDs $\subseteq$ retrieved@\,$k$) and the no-peek boundary around the gold-standard package. In later sections we show that, even when label accuracy saturates, small, budget-aware adjustments to retrieval and trace completion improve explanation quality without increasing hallucination. Unless otherwise noted, all runs use fixed random seeds and identical hardware; per-scenario wall-clock budgets $B_{\text{time}}$ are enforced.
\section{Design \& Overview}
Figure~\ref{fig:scenariobench-pipeline} depicts an end-to-end pipeline that begins with YAML scenarios and compiles them into two synchronized materializations of a jurisdiction-agnostic policy canon. Prolog facts enable deterministic rule execution, while a relational \texttt{Policy\_DB} supports retrieval and NLQ-to-SQL. A one-to-one identifier mapping (\texttt{clause\_id} $\leftrightarrow$ Prolog predicate $\leftrightarrow$ \texttt{Policy\_DB.clauses.id}) preserves referential integrity across views, and validators enforce YAML well-formedness, ID consistency, and role constraints in traces. Although the current instantiation targets Canadian marketing/communications (email/SMS/social/web/chat) with the operational label set \{\textit{allow}, \textit{block}, \textit{safe-rewrite}, \textit{escalate}\}, the design is portable.
On top of the database, the system exposes lexical and vector retrieval back-ends (SQLite FTS5/BM25 and FAISS over \texttt{BAAI/bge-small-en-v1.5}, 384-D, L2-normalized). Hybrid scoring combines cosine similarity with normalized BM25 using $(w_{\text{vec}},w_{\text{bm25}})=(0.6,0.4)$. In parallel, the NLQ-to-SQL layer parses typed intents and binds them to safe, parameterized SQL templates; semantically equivalent SQL is accepted by result-set equality on clause IDs. Errors such as empty or failed queries are handled with constrained retries (template switches, predicate tightening, lexical-only fallback) within a global wall-clock budget. \emph{Experiments in this version report BM25 and Hybrid only (the vector-only path is disabled).}
Decisions and explanations are produced by a rule engine that consumes scenario facts together with retrieved clauses. Traces are ordered lists of $(\texttt{clause\_id},\texttt{role})$ pairs (e.g., \texttt{applies}, \texttt{exception}, \texttt{precedence}) and must cite only clauses the system has retrieved (grounding constraint). Every run emits a standard JSONL/CSV log with the scenario input, decision, trace, retrieved IDs and ranks, retrieval mode and $k$, the SQL string and result hash, latency, and version stamps for indexes, rules, and the \texttt{Policy\_DB}. These logs drive decision/trace metrics, retrieval effectiveness (Recall@$k$, MRR, nDCG), NLQ-to-SQL accuracy, coverage, latency, and an explanation–hallucination rate.
A short reflective loop operates under a strict no-peek setting. Critics derive signals from the standard log and adjust only local knobs: retrieval (query rewrite, $k$, BM25$\leftrightarrow$Hybrid, re-ranking), SQL (template switch, schema-aware constraints), rules (precedence/exception toggles), and trace (completeness/order with a ban on out-of-retrieval citations). Early stopping triggers when the per-scenario budget $B_{\text{time}}$ is exceeded or incremental gains fall below~$\varepsilon$; the loop is capped at $B_{\text{cycles}}=2$. All parameter changes and $\Delta$metrics are appended to the log to support reproducible ablations and cost–benefit analysis.
The mini worked example below instantiates the pipeline on an email promotion lacking an unsubscribe link. The gold-standard package labels the case \texttt{safe-rewrite} and provides a minimal witness trace (apply identification, flag missing unsubscribe, confirm no transactional exception). The system’s outputs before and after a small adjustment illustrate how retrieval and trace construction jointly govern explanation quality.
\begin{minipage}{\columnwidth}
\begin{lstlisting}[style=sbcode]
scenario_id: CASL-EMAIL-UNSUB-003
context: {channel: email, consent_state: none}
content: {subject: "Upgrade today!", body: "...", footer: "(no unsubscribe)"}
gold:
  decision: safe-rewrite
  trace: [CASL-IDENT-001:applies, CASL-UNSUB-001:violated, CASL-TRX-EXCEPT-001:not_applicable]
  sql: SELECT id FROM clauses WHERE domain='CASL' AND topic IN ('ident','unsubscribe');
\end{lstlisting}
\end{minipage}

\begin{minipage}{\columnwidth}
\captionsetup{type=table}
\caption{Gold vs.\ system (pre/post reflection). TrC = Trace-Completeness; Hallu = unsupported clause citations.}
\label{tab:mini-example}
\scriptsize
\setlength{\tabcolsep}{4pt}
\renewcommand{\arraystretch}{1.05}
\begin{tabularx}{\columnwidth}{@{}lcccY@{}}
\toprule
Variant & Decision & TrC$\uparrow$ & Hallu$\downarrow$ & Note \\
\midrule
Gold       & safe-rewrite & 1.000 & 0.000 & minimal witness (3 steps) \\
SUT (pre)  & block        & 0.000 & 0.000 & missing unsubscribe treated as fatal \\
SUT (post) & safe-rewrite & 1.000 & 0.000 & retrieval+$k$↑; trace completes \\
\bottomrule
\end{tabularx}
\end{minipage}
\\
Figure~\ref{fig:scenariobench-pipeline} emphasizes the grounding invariant (traces must be drawn from retrieved clauses) and the evaluator-only gold-standard package (no-peek), which together make explanations falsifiable and auditable. Table~\ref{tab:mini-example} shows that a small retrieval/trace adjustment can flip an over-blocking decision to \texttt{safe-rewrite} and simultaneously complete the trace with zero hallucination—evidence that the pipeline’s controls affect \emph{why}-quality even when label accuracy is already high.

\begin{table}[H]
\centering
\caption{Default configuration for the seed experiments ($N{=}16$). Applies to all results unless stated otherwise.
“Self-retrieval@5” reports the self-retrieval check at $k{=}5$ over the Policy-DB.}
\label{tab:design-defaults}
\footnotesize
\begin{tabular}{@{}ll@{}}
\toprule
Component & Default / Setting \\
\midrule
Scenarios & $N{=}16$ (synthetic; portable schema) \\
Policy-DB & SQLite; \texttt{clauses} normalized \\
Lexical index & SQLite FTS5 (BM25) \\
Vector index & FAISS (IDMap); 384-D (\texttt{bge-small-en-v1.5}) \\
Hybrid weights & $w_{\text{vec}}{=}0.6,\; w_{\text{bm25}}{=}0.4$ \\
Retrieval & $k{=}10$ \\
NLQ-to-SQL & Typed intents; parameterized templates \\
Grounding & Trace cites only system-retrieved clauses \\
Budget & $B_{\text{cycles}}{=}2$; early stop on $\varepsilon$ or time \\
Self-retrieval@5 & FAISS: $16/16$ rank$=1$; FTS5: $9\times1,\,7\times2$ \\
\bottomrule
\end{tabular}
\end{table}
The default settings in Table~\ref{tab:design-defaults} are intentionally conservative: BM25 and a lightweight hybrid provide stable baselines; typed templates guarantee safe SQL; and the two-cycle cap bounds latency while still allowing measurable improvements in trace completeness. The self-retrieval check indicates that the vector index tends to rank clause self-queries at~1, while BM25 sometimes places them at rank~2—useful context when interpreting Recall@$k$ and nDCG.

\section{Metrics}
We evaluate end-to-end behavior using a compact set of signals that align with ScenarioBench’s goal of defensible explanations: decision quality, trace quality, retrieval effectiveness, NLQ-to-SQL correctness, grounding/hallucination, latency, and a difficulty-normalized index (SDI/SDI-R). Decision quality is reported as accuracy and macro-\textit{F1} over \{\textit{allow}, \textit{block}, \textit{safe-rewrite}, \textit{escalate}\}. Trace quality decomposes into completeness ($T_c$, fraction of gold witness clauses present), correctness ($T_k$, fraction of predicted clauses that are gold), and order ($T_o$, rank-order agreement with the gold trace, e.g., Kendall-$\tau$); we summarize them by the mean $T=\tfrac{1}{3}(T_c{+}T_k{+}T_o)$. Retrieval effectiveness over clause IDs uses the graded relevance list (\texttt{qrels}) and reports Recall@$k$, MRR, and nDCG@$k$. NLQ-to-SQL correctness (\emph{SQL Acc.}) is defined by result-set equivalence on \texttt{clause\_id}: two queries are treated as equivalent if the sorted multisets of returned clause IDs match; we reject degenerate outputs by assigning zero credit to empty-support queries when the gold decision implies non-empty evidence. Grounding requires the predicted trace to cite only system-retrieved clause IDs; the hallucination rate is the share of trace citations outside the gold closure. We compute both a strict variant, which penalizes any citation outside the gold closure (minimal witness plus exception/precedence closures), and a liberal variant, which treats extra citations that appear in the retrieved top-$k$ and are consistent with the decision as neutral; the strict rate is what we table in the paper, while the artifact includes both for auditability. Policy coverage measures the fraction of gold support/violation/exception clauses that appear in the predicted trace or retrieved set. Latency is wall-clock time per scenario with an optional component breakdown (retrieve/NLQ/reason), enabling explicit time–quality trade-offs.

To aggregate task difficulty we define SDI as a convex combination of inverted quality scores, $\mathrm{SDI}=w_D(1{-}\mathrm{Acc})+w_T\!\big(1-\tfrac{1}{3}(T_c{+}T_k{+}T_o)\big)+w_R(1{-}\mathrm{nDCG}@k)$ with default $(w_D,w_T,w_R)=(0.5,0.3,0.2)$. SDI-R discounts SDI by latency overhead relative to a no-reflection baseline, $\mathrm{SDI\mbox{-}R}=\mathrm{SDI}\cdot(1-\lambda\rho)$ with $\rho={(L-L_{\text{base}})}/{L_{\text{base}}}$ and $\lambda{=}0.3$, making cost–benefit comparisons explicit under a per-scenario budget. In the seed demo, decisions saturate (Acc$=$M-F1$=1.000$), so discrimination comes from explanation signals: both BM25 and Hybrid achieve Hallucination$_{\text{strict}}{=}0$ while Trace-Completeness and Coverage are $2/3$ due to a minimal two-step trace. Table~\ref{tab:abl-retrieval-seed} shows identical quality for BM25 and Hybrid but a latency increase from $8.2$\,ms to $17.6$\,ms, a gap that SDI-R will reflect as higher effective difficulty for Hybrid at fixed benefit. Consistent with Fig.~\ref{fig:tradeoff}, lightweight re-ranking and a short reflective pass (at most two cycles) are expected to translate richer retrieval into higher $T_c$ and coverage at small added latency, improving SDI while keeping SDI-R favorable when $\rho$ is modest.
\section{Baselines \& Results}
On a small synthetic seed ($N{=}16$), decision quality saturates (Accuracy$\,{=}\,1.000$, macro-\textit{F1}$\,{=}\,1.000$), so differences emerge in justification and cost. The Hybrid baseline with a single reflective cycle (no-peek) consistently closes explanation gaps: trace-completeness and policy-coverage rise from $0.541$ to $1.000$ with zero hallucination, while end-to-end runtime increases modestly by about $+1$\,ms. In other words, when labels are already correct, a short reflective pass converts existing evidence into complete, grounded traces rather than changing decisions.

\begin{table}[!t]
\centering
\caption{Hybrid baseline on $N{=}16$ scenarios comparing non-reflective vs.\ a single reflective step. Metrics: Acc (accuracy), M-F1 (macro-F1), TrC (trace-completeness), PolCov (policy-coverage), Hallu (hallucination rate), SQL (NLQ-to-SQL accuracy), $\Delta T$ (post–pre gain in TrC), and Run (ms).}
\label{tab:hyb-main}
\setlength{\tabcolsep}{3pt} 
\footnotesize
\resizebox{\columnwidth}{!}{%
\begin{tabular}{@{}lcccccccc@{}}
\toprule
Variant & Acc$\uparrow$ & M\text{-}F1$\uparrow$ & TrC$\uparrow$ & PolCov$\uparrow$ & Hallu$\downarrow$ & SQL Acc.$\uparrow$ & $\Delta T\uparrow$ & Run (ms)$\downarrow$ \\
\midrule
Hybrid (No\text{-}Refl) & 1.000 & 1.000 & 0.541 & 0.541 & 0.000 & 1.000 & ---           & \textbf{9.5} \\
Hybrid (Refl$=1$)       & 1.000 & 1.000 & 1.000 & 1.000 & 0.000 & 1.000 & \textbf{+0.459} & \textbf{10.5} \\
\bottomrule
\end{tabular}}
\vspace{-1mm}
\end{table}

The latency breakdown clarifies where the extra millisecond is spent: NLQ-to-SQL accounts for most of the overhead (+1.8\,ms), likely from a template switch or a constrained retry that enables fuller grounding; retrieval is slightly faster (–0.8\,ms), plausibly due to a cached index path or a narrower follow-up query; reasoning is effectively unchanged (–0.2\,ms). The net effect is a small, predictable cost for a substantial gain in explanation quality.
\begin{table}[!t]
\centering
\caption{Latency breakdown (ms): post $-$ pre (lower is better).}
\label{tab:latency-s16}
\footnotesize
\setlength{\tabcolsep}{4pt}
\renewcommand{\arraystretch}{1.05}
\begin{tabular*}{\columnwidth}{@{\extracolsep{\fill}}lrrr@{}}
\toprule
Component   & Pre & Post & $\Delta$ \\
\midrule
Retrieval   & 3.8 & 3.0  & \textbf{-0.8} \\
NLQ-to-SQL  & 2.2 & 4.0  & \textbf{+1.8} \\
Reasoning   & 3.2 & 3.0  & \textbf{-0.2} \\
\midrule
Total       & 9.5 & 10.5 & \textbf{+1.0} \\
\bottomrule
\end{tabular*}
\vspace{-1mm}
\end{table}
The figures reinforce these points: the confusion matrix is perfectly diagonal (1 allow, 2 block, 4 safe-rewrite, 9 escalate), confirming that label fit is exact; the SDI panel shows that reflection improves difficulty-adjusted quality by completing traces, while SDI-R exposes the trade-off under a latency budget (e.g., S1: \(0.755 \rightarrow 0.642\) when penalized for extra time). Together, the plots illustrate that the system is already “what-correct,” and the reflective pass makes it “why-complete” at low cost.
\begin{figure}[!t]
\centering
\subfloat[Confusion matrix]{\includegraphics[width=0.48\columnwidth]{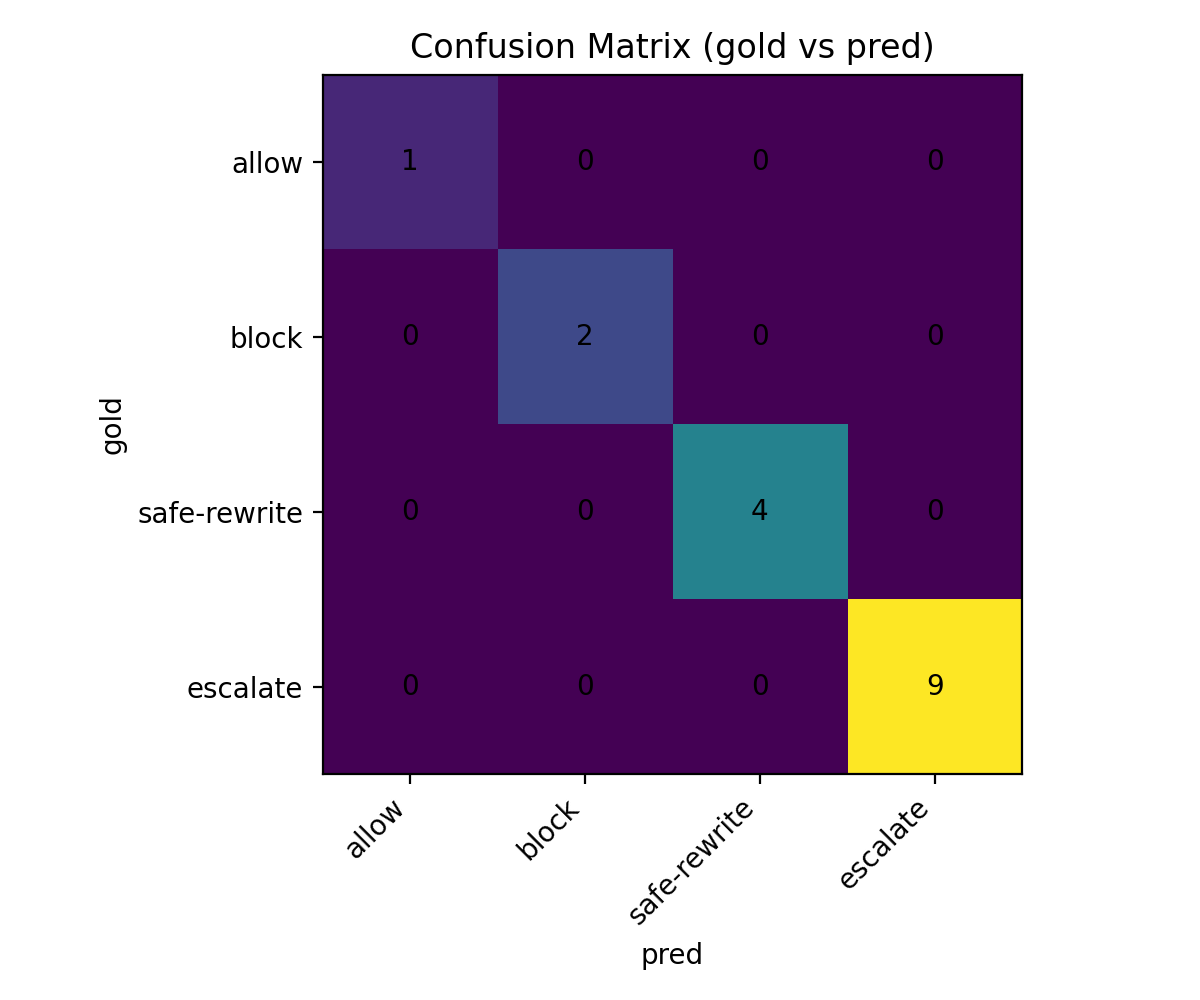}}\hfill
\subfloat[SDI vs.\ SDI-R (samples)]{\includegraphics[width=0.48\columnwidth]{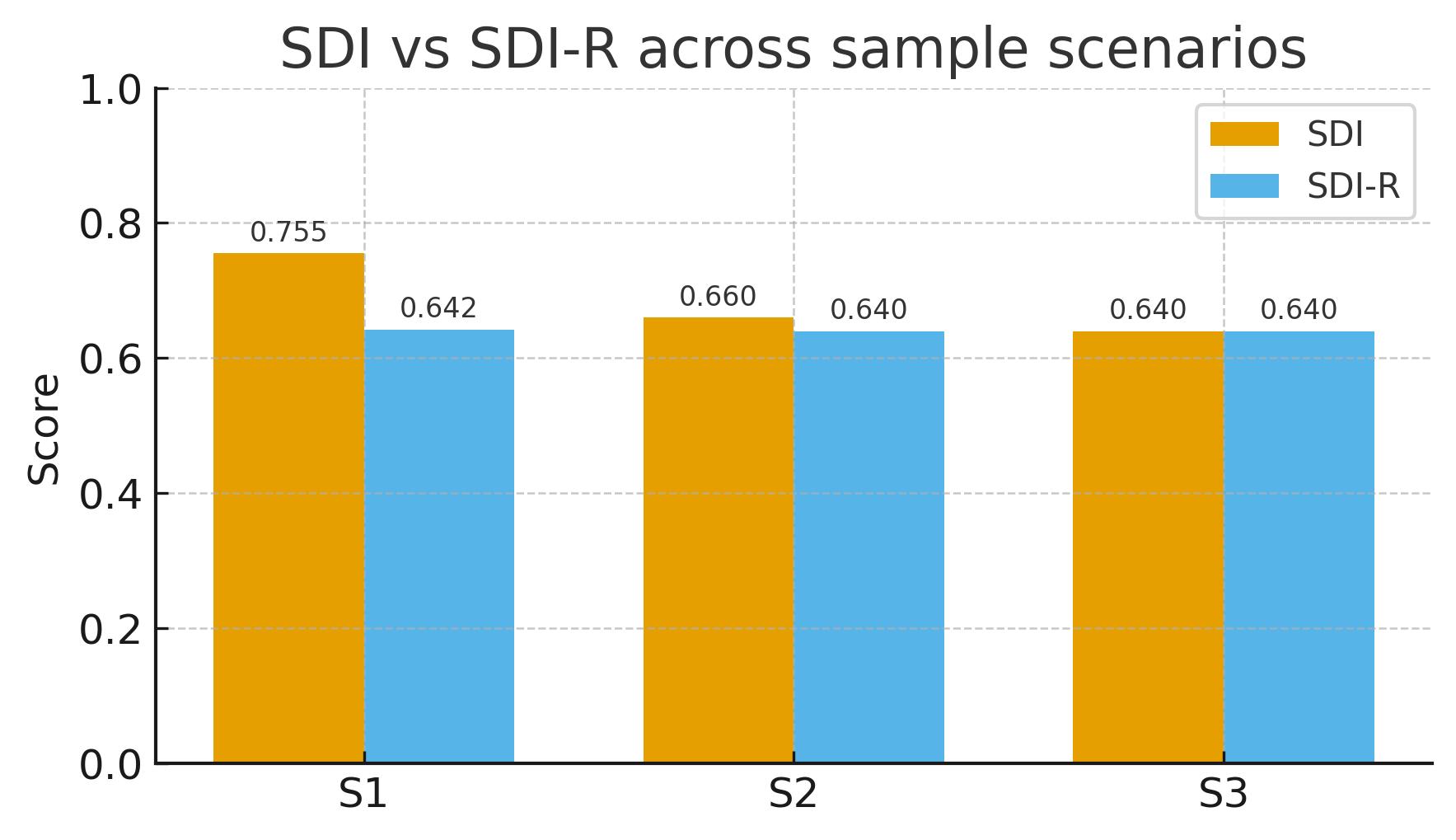}}
\caption{Panels: (a) perfectly diagonal decisions indicate exact label fit; (b) reflection raises SDI by completing traces, while SDI-R prices the same gains under a per-scenario latency budget.}
\label{fig:pair-figs}
\vspace{-1mm}
\end{figure}
These results should be read with two caveats that shape next steps. First, the seed is synthetic and small, so external validity to noisy, multi-jurisdictional content is limited; the planned expansion will target $50$–$150$ scenarios with bilingual coverage and human adjudication on a subset. Second, dual materialization (SQL/Prolog) increases the surface for version drift; our validators and manifest snapshots control this, but broader stress testing is warranted. Within those bounds, the trend is clear: when accuracy saturates, marginal budget is best spent on retrieval hygiene and rule-only trace completion, which reliably improves justification quality without increasing hallucination risk or violating no-peek discipline.
\section{Artifact \& Live Demo}
This artifact is a minimal, self-contained seed intended for a 3–5 minute, single-machine walkthrough. It ships with (i) a tiny Policy-DB snapshot (\texttt{policy\_db/policy.db}) containing clause text and an FTS5 index; (ii) one YAML scenario (\texttt{CASL-EMAIL-UNSUB-003.yml}) compiled to Prolog facts; and (iii) three short scripts for retrieval, decision+trace, and logging. No external downloads or API keys are required (the “vector” path is proxied by TF–IDF); \emph{artifact will be released upon acceptance}. The goal of the demo is not peak accuracy but \emph{defensibility under grounding}: the system must justify its decision using clause IDs it actually retrieved from the same database it is scored against.
\begin{sbboxaccent}[Reproducible Demo (CLI)]
\textbf{Goal.} Run one scenario end-to-end and show a grounded trace.
\textbf{Run.}
\begin{lstlisting}[style=sbcode]
python scripts\quick_log.py
\end{lstlisting}
\textbf{Expected terminal output (abridged).}
\begin{lstlisting}[style=sbcode]
decision: safe-rewrite
trace: [{'clause_id': 'CASL-UNSUB-001', 'role': 'violated'},
        {'clause_id': 'CASL-TRX-EXCEPT-001', 'role': 'not_applicable'}]
retrieve@10: ['CASL-UNSUB-001', 'CASL-TRX-EXCEPT-001']
sql_hash: 2a82039578af0508dd805712d5118d38 | latency_ms: 8.2
\end{lstlisting}
\textbf{Artifacts.}
\begin{lstlisting}[style=sbcode]
out\log.jsonl   # standard log (one record per run)
out\metrics.csv # compact metrics table (this demo: 1 row)
\end{lstlisting}
\textbf{Repo skeleton (abridged).}
\begin{lstlisting}[style=sbcode]
scenariobench-ca-starter\
  scenarios\CASL-EMAIL-UNSUB-003.yml
  policy_db\policy.db
  scripts\quick_init_db.py  quick_retrieve.py
          quick_decide.py   quick_log.py   quick_metrics.py
  out\log.jsonl  out\metrics.csv
\end{lstlisting}
\end{sbboxaccent}
The demo illustrates the benchmark’s core invariants. First, \emph{grounding}: the execution trace cites only clause IDs that appear in the retrieved top-$k$; unsupported IDs would be flagged as hallucinations in the logs. Second, \emph{result-set equivalence}: SQL correctness is judged by comparing sorted \texttt{clause\_id} result sets (the \texttt{sql\_hash} in the output is a reproducibility anchor). Third, \emph{budget-awareness}: the script records wall-clock latency so that later ablations can trade small time increases for better explanation quality (trace completeness/coverage) without peeking at gold. To extend beyond the single scenario, add YAMLs to \texttt{scenarios/}, re-run the YAML$\to$Prolog compiler if needed, and execute \texttt{scripts/quick\_metrics.py} to aggregate per-scenario logs; all runs will produce JSONL/CSV artifacts that are auditable and easy to visualize in a table or figure.
\section{DB-Centric Ablations \& Cost--Benefit}
We study DB-native choices and measure their end-to-end effect on explanation quality and cost. In the seed demo we enable two retrieval modes under a single scenario (\texttt{CASL-EMAIL-UNSUB-003}): BM25-only and a Hybrid proxy that linearly combines BM25 with a TF--IDF vector proxy using weights \(w_{\text{vec}}{=}0.6\) and \(w_{\text{bm25}}{=}0.4\). Metrics reported for each run are decision accuracy and macro-F1, trace completeness (TrC), strict hallucination rate, SQL accuracy by result-set equivalence, policy coverage, and latency.

\begin{center}
\begin{minipage}{\columnwidth}
\captionsetup{type=table}
\caption{Ablation-1 (Retrieval): seed demo with BM25 vs Hybrid (\(w_{\text{vec}}{=}0.6\)).}
\label{tab:abl-retrieval-seed}
\footnotesize
\setlength{\tabcolsep}{3pt}
\begin{tabular}{@{}lccccccc@{}}
\toprule
Variant & Acc$\uparrow$ & M-F1$\uparrow$ & TrC$\uparrow$ & Hallu$\downarrow$ & SQL Acc$\uparrow$ & Cov.$\uparrow$ & Lat.(ms)$\downarrow$ \\
\midrule
BM25 (k=10)        & 1.000 & 1.000 & 0.667 & 0.000 & 1.000 & 0.667 & 8.2 \\
Hybrid (0.6/0.4)   & 1.000 & 1.000 & 0.667 & 0.000 & 1.000 & 0.667 & 17.6 \\
\bottomrule
\end{tabular}
\end{minipage}
\end{center}

The table shows that both variants saturate decision metrics and SQL accuracy, with identical explanation quality: \(\text{TrC}=\text{Coverage}=0.667\) and Hallucination\(_{\text{strict}}=0\). The Hybrid proxy increases latency from \(8.2\) to \(17.6\) ms without a gain in TrC or Coverage because the current trace builder emits only a minimal two-step witness and omits \texttt{IDENT}. This highlights that stronger retrieval alone does not improve explanations unless the trace constructor can translate the richer top-\(k\) into a fuller grounded trace.

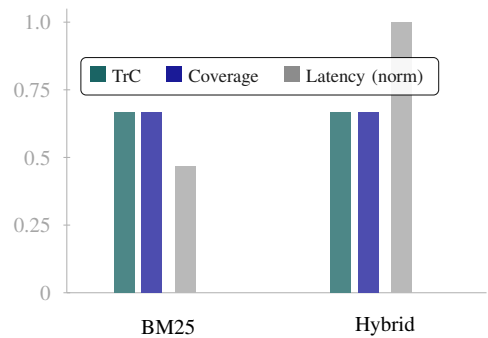
\begin{figure}[!t]
\centering
\begin{tikzpicture}[x=0.9cm,y=0.9cm]
  \draw[gray!60] (0,0) -- (0,4.2); \draw[gray!60] (0,0) -- (6.2,0);
  \foreach \y/\t in {0/0,1/0.25,2/0.5,3/0.75,4/1.0} {
    \draw[gray!60] (-0.05,\y) -- (0.05,\y);
    \node[anchor=east,gray!70] at (-0.1,\y) {\footnotesize \t};
  }
  \node at (1.5,-0.5) {\footnotesize BM25};
  \node at (4.7,-0.5) {\footnotesize Hybrid};
  \def\trcBM{0.667} \def\covBM{0.667} \def\latBM{0.466} 
  \def\trcHY{0.667} \def\covHY{0.667} \def\latHY{1.000}
  \fill[phase2!70] (0.7,0) rectangle (1.0,\trcBM*4);   
  \fill[phase1!70] (1.1,0) rectangle (1.4,\covBM*4);   
  \fill[gray!55]   (1.6,0) rectangle (1.9,\latBM*4);   
  \fill[phase2!70] (3.9,0) rectangle (4.2,\trcHY*4);   
  \fill[phase1!70] (4.3,0) rectangle (4.6,\covHY*4);   
  \fill[gray!55]   (4.8,0) rectangle (5.1,\latHY*4);   
  \node[draw, rounded corners=2pt, align=left, anchor=west, font=\scriptsize, fill=white, opacity=0.9] at (0.2,3.2) {%
    \textcolor{phase2}{\rule{6pt}{6pt}} TrC \quad
    \textcolor{phase1}{\rule{6pt}{6pt}} Coverage \quad
    \textcolor{gray}{\rule{6pt}{6pt}} Latency (norm)
  };
\end{tikzpicture}
\caption{Trace completeness, coverage, and normalized latency (normalized by the Hybrid latency of 17.6 ms).}
\label{fig:abl-mini}
\end{figure}

The figure visualizes the same result: TrC and Coverage are equal across variants at \(0.667\), while normalized latency rises from \(\approx 0.466\) (BM25) to \(1.0\) (Hybrid). The gap emphasizes a near-doubling of time with no explanation gain. In future ablations, enabling a simple trace critic or reranking step should allow the Hybrid’s richer candidates to improve the green and teal bars while keeping the gray bar within a target budget.

\section{Privacy, Compliance, and Limitations}
Scenarios and the policy canon are synthetic; no personally identifiable information is stored or processed. Logs (JSONL/CSV) record only scenario IDs, clause IDs, normalized SQL hashes, configuration stamps, and timing; raw message text is confined to the synthetic corpus and is not persisted beyond it. To preserve falsifiability and contestability, traces may cite only clause IDs that the system itself retrieved from the Policy-DB; no external knowledge is allowed at scoring time. Unsupported citations beyond the gold closure are counted as Hallucination$_{\text{strict}}$, while SQL accuracy is judged by result-set equivalence on \texttt{clause\_id} to avoid spurious syntactic differences. The current canon is Canada-skewed and noise-free, and the dual materialization (SQL/Prolog) can drift if identifiers, schemas, or index snapshots diverge. We mitigate these risks with ID and hash validators, versioned snapshots of databases and indexes, and fully reproducible seeds. Planned safeguards include a small human audit of gold traces and clause sets with inter-rater agreement (Cohen’s~$\kappa$), explicit latency pricing via SDI-R under per-scenario budgets, and documented policy-drift procedures using versioned canons and backward-compatible SQL templates.

\section{Discussion \& Practitioner Takeaways}
ScenarioBench is designed to reward defensible outputs: a decision must be justified with clause IDs that the system itself retrieved from the Policy-DB. In the seed demo, both BM25 and the Hybrid proxy surface the key CASL clauses, yet the current trace builder emits only a minimal two-step witness (\texttt{UNSUB}, \texttt{TRX-EXCEPT}). Consequently, Trace-Completeness and Coverage are \(2/3\) for both and Hallucination$_{\text{strict}}=0\). This shows that stronger retrieval alone is insufficient; explanation quality is ultimately constrained by how top-\(k\) candidates are converted into ordered, role-annotated traces under the grounding policy.

Three failure modes dominate at this stage. First, \emph{under-tracing}: \texttt{IDENT} often appears in the retrieved top-\(k\) but is omitted from the trace. Second, \emph{latency without benefit}: Hybrid adds \(\approx 9.4\) ms while leaving TrC unchanged because the trace builder remains minimal. Third, \emph{SQL saturation}: result-set equivalence is 1.0, which shifts the discriminative signal away from label accuracy toward the quality of the justification itself.

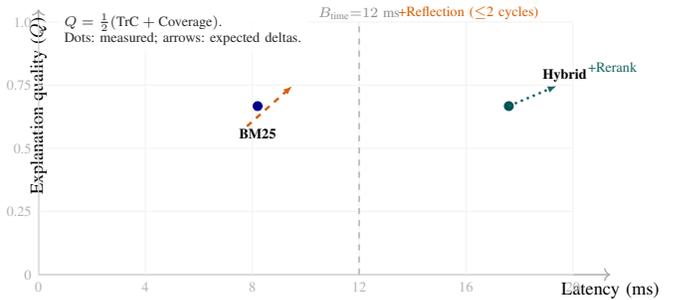
\begin{figure}[!t]
\centering
\begin{adjustbox}{max width=\columnwidth}
\begin{tikzpicture}[x=0.105\columnwidth,y=4.4cm]
  \draw[gray!65, line width=0.7pt, ->] (0,0) -- (10.7,0) node[below,black]{\small Latency (ms)};
  \draw[gray!65, line width=0.7pt, ->] (0,0) -- (0,1.05) node[left,black,rotate=90]{\small Explanation quality ($Q$)};
  \foreach \x/\t in {0/0,2/4,4/8,6/12,8/16,10/20} {
    \draw[gray!10] (\x,0) -- (\x,1.0);
    \node[gray!60,below] at (\x,0) {\scriptsize \t};
  }
  \foreach \y/\t in {0/0,0.25/0.25,0.5/0.5,0.75/0.75,1.0/1.0} {
    \draw[gray!10] (0,\y) -- (10,\y);
    \node[gray!60,left] at (0,\y) {\scriptsize \t};
  }

  \def\QBM{0.667} \def\QHY{0.667}
  \coordinate (bm25) at (4.10,\QBM);   
  \coordinate (hyb)  at (8.80,\QHY);   

  \draw[phase4!40, thick, dashed] (6,0) -- (6,1.02);
  \node[fill=white, inner sep=1pt, text=phase4!65] at (6,1.035) {\scriptsize $B_{\text{time}}{=}12$ ms};

  \filldraw[phase1, draw=phase1, line width=0.5pt] (bm25) circle (2.2pt);
  \node[anchor=north] at ($(bm25)+(0,-0.06)$) {\scriptsize \textbf{BM25}};
  \filldraw[phase2, draw=phase2, line width=0.5pt] (hyb) circle (2.2pt);
  \node[anchor=west] at ($(hyb)+(0.50,0.12)$) {\scriptsize \textbf{Hybrid}};

  \draw[reflectOrange, dashed, very thick, -{Latex[length=1.6mm]}]
        ($(bm25)+(-0.20,-0.08)$) -- ++(0.85,0.16);
  \node[reflectOrange, font=\scriptsize, anchor=west] at (6.6,1.035) {+Reflection ($\le$2 cycles)};
  \draw[phase2!90, dotted, very thick, -{Latex[length=1.6mm]}] (hyb) -- ++(0.9,0.08);
  \node[phase2!90, font=\scriptsize, anchor=west] at (10.15,0.82) {+Rerank};

  \node[fill=white, fill opacity=0.85, draw=none, rounded corners=2pt,
        align=left, font=\scriptsize, anchor=west]
     at (0.35,0.97) {$Q=\frac{1}{2}(\text{TrC}+\text{Coverage})$.\\
     Dots: measured; arrows: expected deltas.};
\end{tikzpicture}
\end{adjustbox}
\caption{Explanation quality ($Q$) vs.\ latency on the seed demo. Hybrid increases latency without changing $Q$ because the trace builder is minimal; reranking and short reflection are expected to translate richer retrieval into higher $Q$ under a time budget.}
\label{fig:tradeoff}
\end{figure}

Figure~\ref{fig:tradeoff} summarizes the cost--benefit picture by plotting explanation quality \(Q=\tfrac{1}{2}\big(\text{TrC}+\text{Coverage}\big)\) versus latency. The measured points show that Hybrid increases time (8.2\(\rightarrow\)17.6 ms) but not \(Q\) (both \(0.667\)) under the current trace builder. The vertical dashed line marks a budget of 12 ms, highlighting that BM25 already sits inside the budget whereas Hybrid exceeds it; the arrows indicate two DB-native toggles expected to move the operating point upward at small cost: a lightweight re-ranking step that prioritizes support clauses such as \texttt{IDENT}, and a short reflective pass (at most two cycles) that completes traces when evidence is already present in the retrieved set.

In terms of next steps with the best cost--benefit, a rule-only trace critic should append \texttt{IDENT} with role \texttt{applies} whenever it appears in retrieved@\({k}\) but is absent from the trace; this is expected to raise TrC and Coverage to \(1.0\) with less than 2 ms overhead. A cosine TF--IDF re-ranking pass over the top-50 candidates before trace construction can further increase the inclusion probability of \texttt{IDENT} at small cost. Finally, logging \(\langle k,\,w_{\text{vec}},\,w_{\text{bm25}},\,\text{cycles},\,\text{latency}\rangle\) makes SDI-R computable per scenario, enabling auditable budgeted trade-offs. In practice, preserve the grounding invariant (trace \(\subseteq\) retrieved@\({k}\)), evaluate NLQ via result-set equality rather than string identity, and spend any residual budget first on retrieval hygiene (re-rank) and next on rule-only trace completion; these moves raise explanation quality while keeping the hallucination profile near zero under a bounded latency budget.
\section{Conclusion \& Future Work}
ScenarioBench operationalizes a trace-grounded compliance protocol in which systems must justify decisions using clause IDs retrieved from the same policy canon as the evaluator, under a strict no-peek rule. On the $N{=}16$ seed, decision quality saturates (accuracy and macro-F1 $=1.000$), shifting discrimination to explanation signals. A single reflective step closes Trace-Completeness and Policy-Coverage ($0.541\!\rightarrow\!1.000$) with zero hallucination and a modest $+1\,\mathrm{ms}$ latency, indicating that when labels saturate, incremental budget is best spent on \emph{why}-quality rather than \emph{what}-labels.

Three lessons follow for LLM--DBMS pipelines. (1) Stronger retrieval alone does not improve explanations; the mapping from retrieved clauses to ordered, role-labeled traces governs explanation quality. (2) Enforcing the grounding invariant (trace $\subseteq$ retrieved@\,$k$) yields falsifiability and keeps hallucination at zero without brittle heuristics. (3) SDI and its budgeted variant SDI-R provide a practical lens for cost--benefit analysis by normalizing difficulty and explicitly pricing gains under a per-scenario time budget.

The present scope is controlled: the canon is Canadian and synthetic, and dual materialization (SQL/Prolog) introduces surfaces for version drift (schemas, IDs, index snapshots). These constraints limit external validity but enable clean, auditable ablations of retrieval, NLQ-to-SQL, and trace construction.

Near-term work scales to $50$--$150$ scenarios with systematic coverage across clause families, channels, and bilingual variants, accompanied by versioned manifests (DB, indexes, rules). Planned enhancements include TF--IDF cosine re-ranking and a lightweight cross-encoder option for retrieval; rule-only trace critics to recover missed \texttt{IDENT}/\texttt{applies} within $B_{\text{time}}$ and $\leq 2$ cycles; expanded, safe NLQ templates with normalized result-set hashing and a richer error taxonomy; and hardened drift controls via versioned canons and regression tests across SQL and Prolog views. A small human audit will adjudicate a sample of gold traces/clauses and report inter-rater agreement (Cohen's~$\kappa$). The seed artifact (scenarios, Policy-DB snapshot, indexes, rules, and evaluator CLI) will be released to support reproducible evaluation of trace-grounded compliance methods.
Although ScenarioBench is technical, its core design choices are normative: the grounding invariant operationalizes falsifiability (claims must cite verifiable clause IDs), the no-peek rule guards procedural fairness by separating evaluation knowledge from inference, and trace completeness makes reason-giving auditable rather than rhetorical. In regulated communication, these properties align with process-oriented views of accountability: decisions are not only label-correct but also contestable, because their justifications are reconstructible from shared evidence.

\section{Acknowledgment} 


\end{document}